\documentclass[letterpaper]{article} \usepackage[preprint]{aaai2027}

\usepackage[hyphens]{url}  \usepackage{graphicx}    \usepackage{natbib}  \usepackage{caption}   \usepackage{amsmath}
\usepackage{amssymb}
\usepackage{tikz}
\usetikzlibrary{arrows.meta,positioning,fit,calc}

\usepackage{algorithm}
\usepackage{algorithmic}

\usepackage{newfloat}
\usepackage{listings}
\DeclareCaptionStyle{ruled}{labelfont=normalfont,labelsep=colon,strut=off} 
\floatstyle{ruled}
\newfloat{listing}{tb}{lst}{}
\floatname{listing}{Listing}

\usepackage{booktabs}

\providecommand{\tightlist}{  \setlength{\itemsep}{0pt}\setlength{\parskip}{0pt}}

\title{Advantage Scale Calibration Imbalance in Group-Relative Optimization under Low-Variance Rewards: Diagnosis and Bounded Recovery}
\author{
    Fei Ding\textsuperscript{\rm 1}\thanks{Corresponding author: Fei Ding (\texttt{dignfei@gmail.com}).},
    Runhao Liu\textsuperscript{\rm 1},
    Yongkang Zhang\textsuperscript{\rm 1},\\
    Yuhao Liao\textsuperscript{\rm 2},
    Zijian Zeng\textsuperscript{\rm 2},
    Huiming Yang\textsuperscript{\rm 2}
}
\affiliations{
    \textsuperscript{\rm 1}Alibaba Group\\
    \textsuperscript{\rm 2}Tsinghua University
}

\begin{document}

\maketitle

\begin{abstract}
In verifier-style RLVR, group-relative optimization often treats advantage scale as an implementation detail. This paper separates two low-variance cases: sub-resolution jitter that should not become a preference signal, and credible but small cardinal gaps that should be learned without distorting KL calibration. We propose an advantage-scale three-way calibration interface: the same within-group scale denominator simultaneously determines the reward-branch strength, prompt-level batch weight, and the effective KL calibration induced when the reward branch is re-expressed on the original cardinal scale. This interface explains why RLOO / Dr.GRPO can let credible small gaps become KL dominated, whereas GRPO's standard-deviation denominator can amplify tiny gaps without bound. Based on this interface, we further introduce the Reward-Resolution Protocol and MaxNorm-AC, respectively filtering sub-resolution gaps and providing bounded cardinal recovery on credible nonzero gaps. Across dense / MoE architectures and math / code reasoning, MaxNorm-AC improves over the prespecified percentile-scale reference (\(p=90\), hereafter the \(p90\) reference) while truncating the low-variance inverse-scale tail.
\end{abstract}

\begin{figure*}[t!]
\centering
\includegraphics[width=0.98\textwidth]{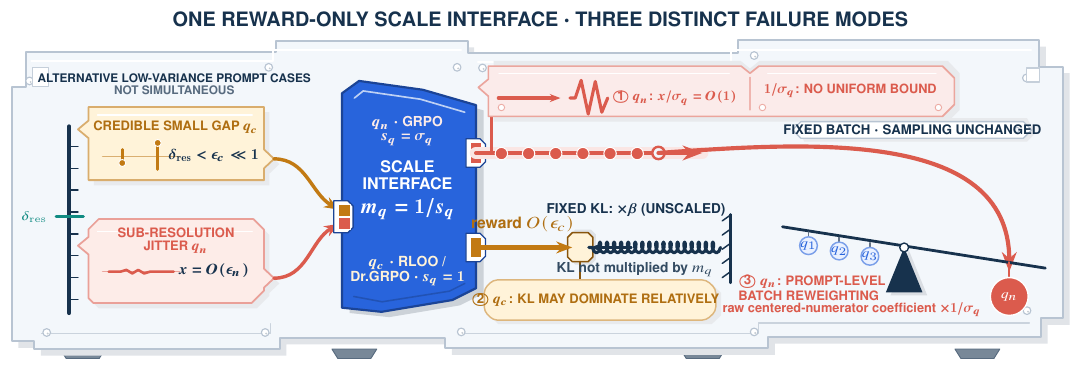}
\caption{Two low-variance regimes expose three existing failures at one reward-scale interface. For sub-resolution jitter \(q_n\), GRPO restores \(x=O(\sigma_q)\) to \(O(1)\), but \(1/\sigma_q\) has no uniform bound and reweights prompts in a fixed batch. For a credible small gap \(q_c\), the \(O(\epsilon_c)\) reward push of RLOO / Dr.GRPO may be relatively dominated by fixed KL. The scale path, spring, and balance encode unbounded amplification, KL pressure, and batch reweighting; our method is not shown.}
\label{fig:summary-mechanism}
\end{figure*}

\section{Introduction}\label{sec:introduction}

In verifier-style RLVR training for reasoning, group-relative optimization constructs advantages from multiple responses to the same prompt and avoids an explicit value function. We distinguish two low-variance cases. A sub-resolution gap is numerical jitter below the minimum credible reward resolution and should not be used as a preference signal. A credible low-variance gap can arise from partial credit, process quality \citep{wang2026grpovpsenhancinggrouprelative,yue2025promotingefficientreasoningverifiable}, confidence, or error severity; such gaps should be learned, but their scale must be calibrated. The Reward-Resolution Protocol first merges sub-resolution gaps before any group-relative update and skips both reward and KL updates for zero-gap groups.

After this protocol, we focus on credible within-group reward gaps that are small but nonzero. In this regime, the prompt-level scale is a three-way interface: it sets the reward-branch strength, the KL calibration seen by that prompt, and the prompt's weight inside the batch. Standard-deviation normalization is therefore not ordinary preprocessing. Without scale recovery, RLOO / Dr.GRPO can become dominated by KL as the reward signal weakens; with standard-deviation recovery, GRPO restores the signal but introduces unbounded amplification, effective-KL imbalance, and prompt-level reweighting as the within-group standard deviation approaches zero. Figure~\ref{fig:summary-mechanism} maps these three existing failures to the drive shaft, fixed-KL spring, and batch balance of one scale machine; our bounded recovery is shown separately in Figure~\ref{fig:maxnorm-pipeline}.

This paper identifies low-variance group-relative updates as a three-way scale-calibration interface and gives a minimal bounded cardinal operator that only replaces the within-group scale denominator. This minimal intervention leaves KL regularization, the policy ratio, clipping, token aggregation, and length normalization unchanged, making the mechanism easier to attribute, the evidence more auditable, and practical transfer lower-cost. MaxNorm-AC replaces the standard-deviation denominator with the maximum absolute raw advantage under a reward-resolution floor and freezes the result as the loss weight for the current optimization step, thereby jointly bounding reward amplification, the effective-KL lower bound, and the prompt-level weight tail; the exact bounds are summarized in the theory section below. The main instance, MaxNorm-RLOO, is only one way to attach this bounded-scale operator. If reward magnitudes are untrustworthy and only the within-group order is reliable, the applicability assumption of our method no longer holds; order-only methods should be treated as external boundaries rather than objects whose superiority is established by this paper. Exactly identical within-group rewards are likewise a signal-construction boundary rather than a scale-calibration case.

The contributions are:
\begin{itemize}
\tightlist
\item
  To our knowledge, we first formulate the within-group denominator in RLVR as a unified scale-calibration problem with three failure modes: \(O(1)\) amplification of sub-resolution jitter, \(O(\epsilon)\) reward-branch decay for credible small nonzero gaps, and prompt-level batch reweighting induced by \(1/s_q\); the denominator should therefore jointly preserve credible cardinal gaps, suppress sub-resolution jitter, and control KL calibration and prompt-level reweighting;
\item
  We propose the Reward-Resolution Protocol, which separates sub-resolution noise from credible cardinal gaps before all group-relative updates through bounded binning, a dead zone, and zero-gap skipping;
\item
  We propose KL-calibrated bounded scale as a diagnostic criterion and use MaxNorm-AC as the minimal bounded intervention that verifies it; the operator only replaces the within-group scale denominator and keeps KL regularization, the policy ratio, clipping, token aggregation, and length normalization unchanged;
\item
  We provide real benchmarks covering dense / MoE architectures and math / code RLVR, compare against the prespecified \(p90\) reference, MAD, Huber, GRPO+\(\beta_q\)-comp, and REINFORCE++, and validate the mechanism through controlled sub-resolution diagnostics, inverse scales, the reward/KL gradient ratio, direction cosine, and KL P95.
\end{itemize}

\section{Background and Related Work}\label{sec:background-related-work}

This paper studies value-free group-relative policy optimization. RLOO uses a leave-one-out REINFORCE baseline \citep{ahmadian-etal-2024-back}; GRPO normalizes centered rewards by the within-group standard deviation \citep{shao2024deepseekmathpushinglimitsmathematical}; Dr.GRPO removes that standard-deviation normalization and fixes length normalization \citep{liu2025understandingr1zeroliketrainingcritical}; REINFORCE++ uses global advantage normalization \citep{hu2025reinforcestabilizingcriticfreepolicy}; and DAPO scales GRPO-style training through system-level recipes such as dynamic sampling and token-level losses \citep{yu2025dapoopensourcellmreinforcement}. These lines make the advantage denominator a central algorithmic choice rather than bookkeeping.

Related work studies difficulty bias, group weighting, KL placement, homogeneous rewards, reward corruption, zero-variance signal construction, trajectory-level correction, hard-example selection, and multi-objective / negative-example pipelines \citep{fontana2026hiddenobjectivebiasesgroupbased,yao2026futureklregularizedgrpoprocesslevel,liu2025rethinkingklregularizationrlhf,he2026advantagecollapsegrouprelative,mansouri2026noisecorrectedgrponoisyrewards,le2026promptleftbehindexploiting,pang2026ticgrpoprovableefficientoptimization,pikus2025hardexamplesneedmaximizing,li2025optimizingsafealignedlanguage,Liu_2026}. Recent GRPO design-space studies, including \(\lambda\)-GRPO, Hybrid GRPO, MEML-GRPO, Stepwise / Spectral-style policy optimization, Sharpness-Guided GRPO, and on-/off-policy GRPO analyses, reshape token preferences, rollout construction, supervision, or update sharpness rather than giving a fixed bounded denominator for credible low-variance cardinal gaps \citep{wang2025lambdagrpounifyinggrpoframeworks,sane2025hybridgrouprelativepolicy,jia2025memlgrpoheterogeneousmultiexpertmutual,chen2026stepwiseguidedpolicyoptimization,le2026sharpnessguidedgrouprelativepolicy,mroueh2025revisitinggrouprelativepolicy}. Compute-supervision analyses of RLVR further motivate treating verifier quality, reward reliability, and rollout structure as part of the learning problem \citep{mitsuhashi2026quantifyingempiricalcomputesupervisiontradeoffs}. Ranking/listwise objectives, robust scales, and bounded-ratio updates are important boundaries or baselines \citep{choi2026gopopolicyoptimizationusing,xiao2025bnpobetanormalizationpolicy,zeng2026shrinkingvarianceshrinkagebaselines,ao2026boundedratioreinforcementlearning}. Robust-scale denominators such as the prespecified \(p90\) reference, MAD, Huber, and std-floor can mitigate outlier scales or inverse-scale tails, but they do not simultaneously satisfy our three-way scale-calibration constraints. Our narrower target is \emph{bounded cardinal advantage calibration}: preserve credible reward gaps while explicitly controlling normalized advantages, effective-KL lower bounds, and prompt-weight tails. Detailed positioning is in the supplementary material.
\section{Problem Setting: The Low-Variance Reward Dilemma}\label{sec:problem-setting}

In group-based reinforcement learning for LLMs, a common practice is to sample multiple responses for the same prompt and construct relative advantages from the rewards of responses in the same group. This paper does not study all within-group learning algorithms, but instead focuses on a more specific question: \textbf{in verifier-style RLVR where reward gaps are credible and low-variance but nonzero, should group-relative advantages be divided by the standard deviation, and is the standard deviation an appropriate denominator?}

\paragraph{Unified notation.}
Let \(r_{q,i}\) be the reward of response \(i\) for prompt \(q\), \(\bar r_q\) and \(\sigma_q\) be the group mean and standard deviation, and \(x_{q,i}=r_{q,i}-\bar r_q\) be the centered reward. To avoid overloading notation, we use \(u_{q,i}\) for the raw numerator, \(c_{q,i}=u_{q,i}/s_q\) for the scale-calibrated advantage, and \(w_{q,i}=\operatorname{sg}(c_{q,i})\) for the frozen loss weight:
\[
\begin{aligned}
u_{q,i} &\quad \text{raw numerator},&
c_{q,i}&=u_{q,i}/s_q,&
w_{q,i}&=\operatorname{sg}(c_{q,i}).
\end{aligned}
\]
Only \(w_{q,i}\) enters the policy-gradient loss. GRPO uses \(A^{\mathrm{std}}_{q,i}=x_{q,i}/\sigma_q\), where \(\sigma_q=(G^{-1}\sum_j(r_{q,j}-\bar r_q)^2)^{1/2}\). RLOO uses \(a^{\mathrm{RLOO}}_{q,i}=r_{q,i}-(G-1)^{-1}\sum_{j\ne i}r_{q,j}\). Thus the main question is the scale choice: RLOO / Dr.GRPO preserve raw gaps and can suffer signal depletion, whereas GRPO restores magnitude but introduces a \(1/\sigma_q\) prompt-dependent weight. MaxNorm-AC preserves credible cardinal gaps while bounding this weight; MaxNorm-RLOO applies the operator to the RLOO numerator.

\section{Without Standard-Deviation Division: RLOO / Dr.GRPO and Signal Depletion}\label{sec:no-std-signal-depletion}

RLOO is naturally obtained from REINFORCE with a leave-one-out baseline, not as a GRPO variant. With the group mean including the current sample, \(a^{\mathrm{RLOO}}_{q,i}=\frac{G}{G-1}(r_{q,i}-\bar r_q)\), so the RLOO numerator and the centered GRPO numerator have the same signs, rankings, and directions; the derivation is in the supplementary material. The core distinction is therefore whether the centered signal is divided by \(\sigma_q\). Removing that denominator avoids random prompt reweighting, but under low-variance rewards it can also leave the reward branch too weak.

\section{Consequences of Signal Depletion: KL Domination and Policy Regression}\label{sec:signal-depletion-kl}

If the credible reward gap under the same prompt is small, the raw RLOO / Dr.GRPO advantage also becomes small. For \(r=[0.51,0.50,0.49,0.50]\), the RLOO advantage is approximately \([0.0133,0,-0.0133,0]\). A KL-regularized update can be abstracted as

\begin{equation}
g_{\mathrm{total}}=g_R-\beta g_{\mathrm{KL}}.
\end{equation}

When low variance shrinks \(\|g_R\|\) below \(\beta\|g_{\mathrm{KL}}\|\), the total direction becomes closer to \(-g_{\mathrm{KL}}\), and the update primarily reduces the KL between the current policy and the reference policy rather than moving in the direction that improves task reward. This phenomenon is not an ordinary plateau: if the policy shift away from the reference is precisely the direction that yields reasoning gains, a KL-dominated update can pull the model back toward the reference model, appearing as a decline in accuracy or reward together with a decrease in KL.

Thus, avoiding scale recovery prevents GRPO's denominator explosion but exposes the opposite failure mode: credible low-magnitude gaps may not provide a sufficiently strong reward signal. Gradient-norm bounds and KL-decrease conditions are in the supplementary material.

\section{Equivalent Interpretation of KL Domination and Performance Regression}\label{sec:kl-domination-equivalence}

Let \(g_R\) be the unattenuated reward gradient and \(g_{\mathrm{KL}}\) the KL gradient. If a credible low-variance signal shrinks the reward branch to \(\alpha g_R\), \(0<\alpha<1\), then the total update direction is

\begin{equation}
g_\alpha=\alpha g_R-\beta g_{\mathrm{KL}}
=\alpha\left(g_R-\frac{\beta}{\alpha}g_{\mathrm{KL}}\right).
\end{equation}

Thus, in the sense that an overall positive scaling does not change the direction, low-variance signal attenuation is equivalent to increasing the KL coefficient to

\begin{equation}
\beta_{\mathrm{eff}}=\frac{\beta}{\alpha}.
\end{equation}

If \(\alpha\|g_R\|<\beta\|g_{\mathrm{KL}}\|\), the update is closer to \(-g_{\mathrm{KL}}\) and can reduce KL to the reference while also reducing task performance. We use this only as a local diagnostic; rigorous small-step proofs are in the supplementary material.

\section{Standard-Deviation Normalization: Signal Recovery and Scale Rewriting}\label{sec:std-normalization}

Standard-deviation normalization can recover low-variance signals, but it recovers a recalibrated surrogate rather than the original cardinal reward objective itself. For a single prompt \(q\), let \(g_{R,q}^{0}\) denote the cardinal reward gradient without standard-deviation division, \(g_{\mathrm{KL},q}\) denote the KL gradient, and \(s_q\) denote the within-group scale denominator. A GRPO-style update can be written as

\begin{equation}
g_q^{\mathrm{GRPO}}
=
\frac{1}{s_q}g_{R,q}^{0}-\beta g_{\mathrm{KL},q}
=
\frac{1}{s_q}\left(g_{R,q}^{0}-\beta s_q g_{\mathrm{KL},q}\right).
\end{equation}

Because \(s_q>0\), if the current batch and denominator are frozen and the pre-optimizer first-order direction is viewed on the unnormalized cardinal reward scale, this direction can be diagnostically re-expressed with the local effective KL coefficient

\begin{equation}
\beta_{\mathrm{eff},q}^{\mathrm{GRPO}}=\beta s_q.
\end{equation}

When \(s_q=\sigma_q\to0\), GRPO restores an \(O(s_q)\) reward difference to an \(O(1)\) advantage, while the prompt's local effective KL diagnostic becomes \(\beta\sigma_q\) and its reward-branch batch weight becomes \(1/\sigma_q\). Thus training may enter a local calibration imbalance: low variance can weaken the reward branch and increase the risk of KL-dominated regression, whereas an overly small standard-deviation denominator can amplify reward noise and prompt-level weight fluctuations, reducing overall training stability. This is a diagnosable local mechanism, not a guarantee about the full PPO / AdamW optimization dynamics. We therefore filter sub-resolution gaps first, then apply bounded scale recovery only to nonzero cardinal gaps that pass the protocol. General derivations, batch reweighting, and AdamW caveats are in the supplementary material.
\section{Reward-Resolution Protocol and MaxNorm-AC}\label{sec:maxnorm-ac-main}

\begin{figure*}[t]
\centering
\includegraphics[width=0.96\textwidth]{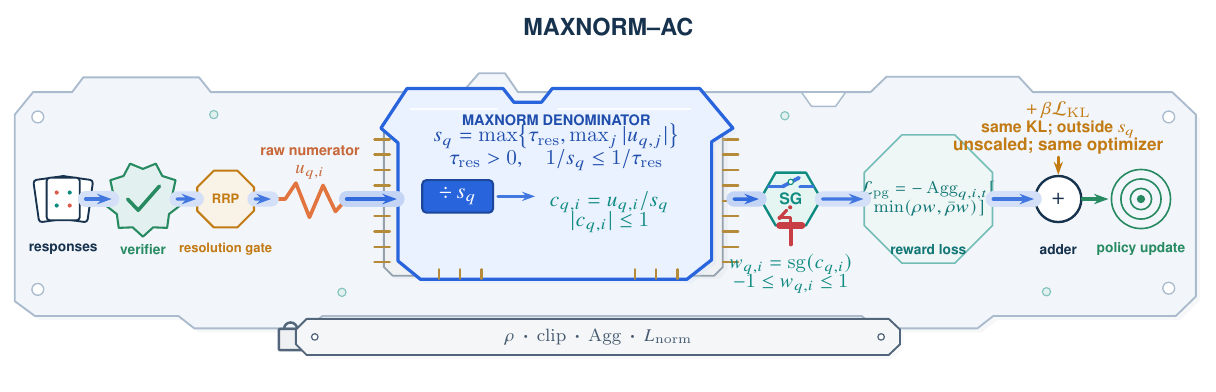}
\caption{MaxNorm-AC inserts the scale card \(s_q=\max\{\tau_{\mathrm{res}},\max_j|u_{q,j}|\}\) only into the denominator of the trusted raw numerator \(u_{q,i}\), yielding ratio- and rank-preserving \(c_{q,i}=u_{q,i}/s_q\) with \(|c_{q,i}|\le1\). The frozen weight \(w_{q,i}=\operatorname{sg}(c_{q,i})\) enters only the reward branch: gradients flow through \(\rho_\theta\) and stop at SG, while base KL joins independently without division by \(s_q\). The probability ratio, clipping, token aggregation, length normalization, and optimizer remain unchanged; the card replaces GRPO's \(\sigma_q\) or the equivalent denominator \(1\) of unscaled methods.}
\label{fig:maxnorm-pipeline}
\end{figure*}

\subsection{Reward-Resolution Protocol}

Arbitrarily small numerical differences in continuous verifier rewards do not necessarily correspond to credible preferences. To prevent group-relative methods from treating sub-resolution noise as preference signal, this paper defines a unified Reward-Resolution Protocol in the reward-function design / reward pipeline:

\begin{itemize}
\tightlist
\item
  \textbf{Bounded rewards}: clip rewards to a fixed range such as \([0,1]\) and cap the contribution of each reward component; single anomalous or extreme reward values should be handled at the reward-function design stage through bounding, clipping, saturating maps, and per-component contribution caps, rather than detected by MaxNorm-AC;
\item
  \textbf{Minimum credible resolution}: set the minimum credible reward resolution to \(\delta_{\mathrm{res}}>0\). Differences below this resolution are treated as belonging to the same bin and do not induce a relative preference;
\item
  \textbf{Zero-gap skipping}: if all responses to the same prompt fall into the same bin, the group contains no credible cardinal reward gap, and both reward and KL updates are skipped during training to avoid KL-only reference regression.
\end{itemize}

This protocol is not an additional optimizer, but a predefined reliability gate that specifies which cardinal reward gaps are credible enough to enter subsequent bounded scale calibration. All methods share bounded rewards, component caps, data, and optimizer settings; the resolution gate, train-time merging, and zero-gap skipping are reported as method-level switches in Table \ref{tab:main-component-attribution}. The standard GRPO row disables the resolution gate, whereas GRPO+Protocol adds only that gate to isolate its effect. In our \([0,1]\) verifier setting, \(\delta_{\mathrm{res}}=10^{-2}\) is fixed from the reward-design resolution before training rather than chosen by validation search; more concretely, if reward component \(k\) has minimum effective step \(\Delta_k\) and aggregation weight \(w_k>0\), then one auditable component-level change contributes at least \(\delta_{\mathrm{pipe}}=\min_{k:w_k>0} w_k\Delta_k\) after aggregation, and we set \(\delta_{\mathrm{res}}\) to the same order as this pipeline resolution as a conservative gate rather than a MaxNorm-AC hyperparameter tuned on validation accuracy. Differences below this scale are treated as unresolved after component quantization, clipping, process-score aggregation, and repeated verifier noise. The \(\tau_{\mathrm{res}}\) used in MaxNorm-AC denotes the explicit safety floor induced by this reward resolution in the advantage denominator and satisfies \(\tau_{\mathrm{res}}=\Theta(\delta_{\mathrm{res}})\); it is not an independently searched optimization hyperparameter, and zero-gap groups are accounted for separately through the reported exact zero-gap skip rate. If a reward pipeline lacks explicit reward-component resolutions, \(\delta_{\mathrm{res}}\) should be pre-calibrated from an upper quantile of reward jitter estimated by repeated verification, expert re-labeling, or historical agreement, with neighboring sensitivity reported rather than validation-accuracy search.

\subsection{MaxNorm-AC}\label{sec:maxnorm-ac}

For a low-variance group that still contains credible nonzero gaps after passing the Reward-Resolution Protocol, our method chooses neither no normalization nor standard-deviation division, but instead uses the maximum absolute raw numerator with a reward-resolution floor as the denominator. We call this module MaxNorm-AC. Its goal is to recover credible low-variance signals while avoiding unbounded signal amplification caused by the standard-deviation denominator, and to bound the arbitrarily large relative weight that a low-variance group can receive in batch aggregation.

Let \(u_{q,i}\) denote the raw numerator produced by any within-group method. Taking RLOO as an example, the raw numerator is:

\begin{equation}
a_{q,i}^{\mathrm{RLOO}}
=
r_{q,i} - \frac{1}{G-1} \sum_{j\ne i} r_{q,j}.
\end{equation}

The centered numerator of GRPO or Dr.GRPO can be written as:

\begin{equation}
x_{q,i}=r_{q,i}-\bar r_q,
\qquad
u_{q,i}=x_{q,i}.
\end{equation}

MaxNorm-AC uniformly uses the maximum absolute raw advantage within the group for bounded recovery, and freezes the scale-calibrated advantage as a fixed weight in the policy-gradient loss:

\begin{equation}
\begin{aligned}
s_q
&=\max\{\max_{1\le j\le G}|u_{q,j}|,\tau_{\mathrm{res}}\},\\
c_{q,i}
&=\frac{u_{q,i}}{s_q},\\
w_{q,i}
&=\operatorname{sg}\left(c_{q,i}\right).
\end{aligned}
\end{equation}

Here \(\tau_{\mathrm{res}}\) is the reward-resolution floor given by the Reward-Resolution Protocol, and \(\operatorname{sg}(\cdot)\) denotes stop-gradient; the denominator is not learned or optimized through backpropagation. Since \(s_q\ge\max_j|u_{q,j}|\), the scale-calibrated advantage satisfies \(|c_{q,i}|\le1\), and the frozen weight entering the loss also satisfies \(|w_{q,i}|\le1\), while preserving the sign and within-group ranking direction of the raw numerator. MaxNorm-AC only replaces the advantage scale denominator; it does not change the policy ratio, clipping, length normalization, or reward construction. It bounds the unbounded reward-branch reweighting that may arise in GRPO as \(\sigma_q\to0\).

When \(u_{q,i}=a_{q,i}^{\mathrm{RLOO}}\), the expression corresponds to MaxNorm-RLOO; when \(u_{q,i}=r_{q,i}-\bar r_q\), it replaces the standard-deviation denominator in GRPO or Dr.GRPO. We report MaxNorm-RLOO as the representative instance because the RLOO and centered numerators differ by a positive constant when the floor is inactive, and we include MaxNorm-Dr.GRPO in Tables \ref{tab:main-per-task-gains} and \ref{tab:main-evidence} to show that the result is not tied to the RLOO numerator. In the example above, \(a^{\mathrm{RLOO}}_{q,1:4}\approx[0.0133,0,-0.0133,0]\) becomes \(c_{q,1:4}\approx[1,0,-1,0]\), while \(1/s_q\le1/\tau_{\mathrm{res}}\) still bounds the prompt's batch weight.

\section{Algorithmic Placement and Training Recipe}\label{sec:algorithm-recipe}

To avoid ambiguity about the implementation location, this paper writes MaxNorm-AC as an independent advantage-calibration operator. Figure~\ref{fig:maxnorm-pipeline} depicts the intervention as a single scale cartridge placed after reward / advantage construction and before the policy-gradient loss; KL regularization, policy ratio, clipping, token aggregation, and length normalization all retain the original definitions of the base algorithm.

\noindent\textbf{Algorithm 1: MaxNorm-AC training steps.}
For each prompt, sample \(G\) responses, compute rewards, construct a raw numerator \(u_{q,i}=B(q,o_{q,i},\{r_{q,j}\}_{j=1}^G)\) from the base group-relative method, set \(s_q=\max\{\max_j|u_{q,j}|,\tau_{\mathrm{res}}\}\), and replace the base loss advantage by \(w_{q,i}=\operatorname{sg}(u_{q,i}/s_q)\).

In Eqs. (8)--(10), \(u_{q,i},s_q,c_{q,i},w_{q,i}\), sampled rollouts, rewards, and \(L_{\mathrm{norm}}\) are fixed for the current update; gradients are taken only with respect to policy parameters \(\theta\) through \(\rho_{q,i,t}(\theta)\), \(\bar\rho_{q,i,t}(\theta)\), and \(\mathcal L_{\mathrm{KL}}(\theta)\). For a PPO / GRPO-style clipped loss with ratio \(\rho_{q,i,t}\) and clipped ratio \(\bar\rho_{q,i,t}\), the reward branch becomes

\begin{equation}
\begin{aligned}
\mathcal L_{\mathrm{pg}}(\theta)
&=
-\mathbb E_q
\left[
\frac{1}{G}
\sum_{i=1}^G
\frac{1}{L_{\mathrm{norm}}}
\sum_{t=1}^{|o_{q,i}|}
\ell^{\mathrm{AC}}_{q,i,t}(\theta)
\right],\\
\ell^{\mathrm{AC}}_{q,i,t}(\theta)
&=
\min\{\rho_{q,i,t}(\theta)w_{q,i},
\bar\rho_{q,i,t}(\theta)w_{q,i}\}.
\end{aligned}
\end{equation}

Here \(L_{\mathrm{norm}}\) is inherited from the base algorithm: response-level averaging uses \(|o_{q,i}|\), while Dr.GRPO uses the fixed budget \(L_{\max}\). KL regularization remains outside the reward branch,

\begin{equation}
\mathcal L(\theta)
=
\mathcal L_{\mathrm{pg}}(\theta)
+
\beta \mathcal L_{\mathrm{KL}}(\theta),
\end{equation}

where \(\mathcal L_{\mathrm{KL}}\) can use the token-level or sequence-level KL already provided by the base algorithm. MaxNorm-AC does not apply \(s_q\) to the KL term; it changes reward/KL calibration through the reward branch only. Thus it composes with Dr.GRPO: MaxNorm-AC handles the question-level denominator, while the fixed length budget handles response-level length normalization.

\section{Theory Summary}\label{sec:theory-main-summary}

The main-text theory is a constructive failure-mode analysis plus a bounded-recovery criterion: under a fixed current batch, fixed scale denominators, and a local first-order surrogate, it formalizes sub-resolution jitter amplification, small-gap reward-branch decay, and prompt-level batch reweighting, then gives the bounds that MaxNorm-AC enforces on the same denominator interface. The \(\tau_{\mathrm{res}}>0\) supplied by the Reward-Resolution Protocol defines
\begin{equation}
s_q=\max\{\max_j|u_{q,j}|,\tau_{\mathrm{res}}\}.
\end{equation}
It immediately gives the three bounds used as diagnostics:
\[
\begin{aligned}
|w_{q,i}| &\le 1,\\
1/s_q &\le 1/\tau_{\mathrm{res}},\\
\beta_{\mathrm{eff},q}=\beta s_q &\ge \beta\tau_{\mathrm{res}} .
\end{aligned}
\]
The first two bounds control the normalized advantage and prompt-level amplification; the third is a fixed-batch, fixed-denominator local diagnostic on the original cardinal reward scale, not a claim that PPO/AdamW optimizes with prompt-specific \(\beta s_q\). In the main setting, \(\tau_{\mathrm{res}}=10^{-2}\) and \(\beta_{\mathrm{kl}}=0.002\), giving \(\beta_{\mathrm{eff},q}\ge2\times10^{-5}\) for retained nonzero-gap groups. MaxNorm-AC introduces bounded prompt-level reweighting bias; because the same positive scale is shared within each prompt, it preserves within-group signs, ordering, and relative-magnitude ratios.

\begin{table*}[t!]
\centering
\small
\resizebox{\textwidth}{!}{\begin{tabular}{llcccccc}
\toprule
Model & Benchmark & Strongest core baseline & Prespecified \(p90\) ref. & GOPO & MaxNorm-Dr.GRPO & MaxNorm-RLOO & \(\Delta\text{ vs }p90\text{ ref.}\) \\
\midrule
Qwen3-32B & AIME25 & \(78.8\pm1.3\) & \(79.4\pm0.9\) & \(78.8\pm0.9\) & \(84.0\pm1.0\) & \(\mathbf{84.2\pm1.2}\) & \(+4.8\) \\
Qwen3-32B & HMMT25 & \(58.2\pm1.1\) & \(59.1\pm1.0\) & \(58.4\pm1.1\) & \(62.9\pm0.8\) & \(\mathbf{63.0\pm0.9}\) & \(+3.9\) \\
Qwen3-32B & LiveCodeBench v6 & \(63.4\pm0.8\) & \(63.9\pm0.8\) & \(63.3\pm1.0\) & \(69.4\pm1.1\) & \(\mathbf{69.7\pm0.7}\) & \(+5.8\) \\
Qwen3-Next & AIME25 & \(89.8\pm0.9\) & \(90.0\pm1.2\) & \(89.5\pm0.9\) & \(93.7\pm0.9\) & \(\mathbf{93.7\pm1.1}\) & \(+3.7\) \\
Qwen3-Next & HMMT25 & \(77.1\pm1.0\) & \(77.5\pm1.1\) & \(76.8\pm1.2\) & \(82.1\pm1.2\) & \(\mathbf{82.3\pm1.0}\) & \(+4.8\) \\
Qwen3-Next & LiveCodeBench v6 & \(70.8\pm1.0\) & \(71.0\pm0.7\) & \(70.4\pm0.8\) & \(76.5\pm0.7\) & \(\mathbf{76.6\pm0.9}\) & \(+5.6\) \\
\bottomrule
\end{tabular}}
\caption{Per-task benchmark gains restored in the main text. The strongest core baseline is selected per model--task unit among RLOO + Protocol, GRPO, GRPO+\(\beta_q\)-comp, Dr.GRPO, and REINFORCE++; GRPO+\(\beta_q\)-comp is selected in all six units. The prespecified \(p90\) reference column exposes the percentile-scale paired comparison, with MaxNorm-RLOO higher on all six rows. GOPO is the pure-ranking boundary baseline. MaxNorm-Dr.GRPO is the centered-numerator check; \(\Delta\text{ vs }p90\text{ ref.}\) reports the MaxNorm-RLOO improvement over this reference.}
\label{tab:main-per-task-gains}
\end{table*}

This criterion also clarifies the limits of the prespecified \(p90\) reference, MAD, Huber, and std-floor. The \(p90\) reference, MAD, and Huber are robust scale estimators: they reduce outlier influence, but their denominators need not cover the largest credible numerator and therefore do not constructively guarantee the normalized-advantage bound. std-floor only truncates the inverse-\(\sigma_q\) tail and likewise cannot bound the largest normalized advantage. MaxNorm-AC puts the maximum absolute raw numerator directly into the denominator, so the three bounds hold simultaneously.

The theoretical claim is therefore a testable three-way calibration criterion: for every retained nonzero group, the denominator jointly modulates reward strength, prompt weight, and KL calibration on the original reward scale, and should induce coupled changes in the \(1/s_q\) tail, reward/KL ratio, contribution concentration, direction cosine, and matched-KL ranking; tail clipping without improved alignment or matched-KL performance is insufficient. The supplementary material gives controlled numerical diagnostics, the AdamW scaling derivation, and first-order sufficient conditions for bounded prompt reweighting to remain an improvement direction.
\section{Empirical Results and Core Diagnostics}\label{sec:empirical-results}

\begin{table}[b!]
\centering
\scriptsize
\setlength{\tabcolsep}{1.4pt}
\renewcommand{\arraystretch}{1.02}
\begin{tabular}{@{}p{0.35\columnwidth}ccccc@{}}
\toprule
Configuration & Protocol & \shortstack{train-time\\\(\delta_{\mathrm{res}}\)-bin} & \shortstack{ZG\\skip} & Acc & KL P95 \\
\midrule
GRPO & N & N & Y & \(70.6\) & \(0.086\) \\
GRPO + Protocol & Y & N & Y & \(71.3\) & \(0.091\) \\
RLOO + Protocol & Y & N & Y & \(71.4\) & \(0.032\) \\
\(p90\) ref. + Protocol & Y & N & Y & \(73.5\) & \(0.080\) \\
MaxNorm-RLOO, no \(\delta_{\mathrm{res}}\)-binning & Y & N & Y & \(78.3\) & \(0.055\) \\
MaxNorm-RLOO, Protocol off & N & Y & Y & \(78.3^{\dagger}\) & \(0.055^{\dagger}\) \\
MaxNorm-RLOO, Protocol/binning off & N & N & Y & \(78.0\) & \(0.058\) \\
MaxNorm-RLOO, keep ZG KL & Y & Y & N & \(77.8\) & \(0.041\) \\
MaxNorm-Dr.GRPO & Y & Y & Y & \(78.1\) & \(0.051\) \\
Full MaxNorm-RLOO & Y & Y & Y & \(\mathbf{78.3}\) & \(0.055\) \\
\bottomrule
\end{tabular}
\caption{Single-column component ablation. The Protocol column indicates whether the reward-pipeline resolution gate is enabled; even when Protocol is disabled, reward bounding, component caps, data, and optimizer settings remain fixed. The train-time \(\delta_{\mathrm{res}}\)-binning column only indicates whether the algorithm additionally merges sub-resolution gaps before advantage construction. GRPO and GRPO+Protocol use the same standard-deviation denominator and isolate the resolution gate; RLOO+Protocol and MaxNorm-RLOO no-\(\delta_{\mathrm{res}}\)-binning share the three settings and isolate denominator calibration; \(\dagger\) marks the Protocol-off but train-time-binned row, which has the same effective gaps as Full MaxNorm-RLOO. The Protocol/binning off row reports measured \(78.0/0.058\).}
\label{tab:main-component-attribution}
\end{table}

Tables \ref{tab:main-per-task-gains}--\ref{tab:main-component-attribution} separate benchmark, diagnostic, and ablation evidence. We use Qwen3-32B and Qwen3-Next-80B-A3B-Thinking; train on decontaminated DeepMath-103K and OpenCodeReasoning \citep{he2025deepmath103klargescalechallengingdecontaminated,ahmad2025opencodereasoningadvancingdatadistillation}; and evaluate on AIME25, HMMT25, and LiveCodeBench v6 \citep{maa_aime2025,balunovic2026matharenaevaluatingllmsuncontaminated,jain2024livecodebenchholisticcontaminationfree}. All methods share five seeds, data, sampling budget, optimizer, learning rate, \(\beta_{\mathrm{kl}}\), max length, batch size, \(G=16\), validation prompts, and checkpoint indices; baseline scale hyperparameters use the shared beta audit and prespecified method-specific settings documented in the supplement, with \(p=90\) fixed before training for the \(p90\) reference and \(\tau_{\mathrm{res}}=\delta_{\mathrm{res}}\) fixed outside validation search for MaxNorm-RLOO. Entries are seed means with seed-bootstrap \(95\%\) CIs, and the paired \(p90\)-reference CI bootstraps six model--task differences. Official results use the final same-compute checkpoint; the supplement reports matched-KL and saved-checkpoint audits, a consolidated training and sampling protocol, OOD retention, and validation prompts.

\begin{table*}[t]
\centering
\small
\begin{tabular*}{\textwidth}{@{\extracolsep{\fill}}lccccc@{}}
\toprule
Method / check & mean KL & KL P95 & Inv P95/P99 & R/KL & Dir. \\
\midrule
RLOO + Protocol & \(0.017\) & \(0.032\) & \(1/1\) & \(0.48\) & \(1.00\) \\
GRPO & \(0.044\) & \(0.086\) & \(980/3100\) & \(3.10\) & \(0.61\) \\
GRPO+\(\beta_q\)-comp & \(0.030\) & \(0.061\) & \(980/3100\) & \(2.94\) & \(0.63\) \\
Dr.GRPO & \(0.019\) & \(0.034\) & \(1/1\) & \(0.51\) & \(1.00\) \\
REINFORCE++ & \(0.032\) & \(0.058\) & \(2.2/3.1\) & \(1.38\) & \(0.91\) \\
\(p90\) ref. + Protocol & \(0.036\) & \(0.080\) & \(100/100\) & \(1.95\) & \(0.84\) \\
MaxNorm-Dr.GRPO & \(0.026\) & \(0.051\) & \(100/100\) & -- & -- \\
MaxNorm-RLOO & \(0.028\) & \(0.055\) & \(100/100\) & \(1.72\) & \(0.88\) \\
\bottomrule
\end{tabular*}
\caption{Low-variance diagnostics on updated groups with \(\sigma_q<10^{-2}\). R/KL is \(\|g_R\|/(\beta\|g_{\mathrm{KL}}\|)\), and Dir. is the reward/KL direction cosine. Per-task accuracy and paired \(p90\)-reference comparisons are in Table \ref{tab:main-per-task-gains}; checkpoint audits are in the supplement.}
\label{tab:main-evidence}
\end{table*}

All official methods share the same bounded reward construction, component caps, and training budget; Table \ref{tab:main-component-attribution} reports the Reward-Resolution Protocol, train-time \(\delta_{\mathrm{res}}\)-binning, and exact zero-gap skipping as row-level switches. Standard GRPO disables the resolution gate, whereas GRPO+Protocol enables only that gate; when enabled, \(\delta_{\mathrm{res}}=10^{-2}\) is fixed before training by reward-design resolution rather than validation search, and differences below this scale are treated as unresolved after component quantization, clipping, process-score aggregation, and repeated verifier noise. We audit this fixed gate instead of tuning it: nearby \(\tau_{\mathrm{res}}\in\{5\times10^{-3},10^{-2},2\times10^{-2}\}\) gives gains \(+4.65,+5.23,+4.42\), the main setting has \(1/s_q\le100\) and \(\beta_{\mathrm{eff},q}\ge2\times10^{-5}\), floor activation is \(11.8\%-20.4\%\) of nonzero updated groups, and the exact zero-gap skip rate in the default setting is \(2.4\%\). Thus the evidence is exposed as a fixed reliability gate, local sensitivity check, and transparent exact zero-gap skip rate report, rather than a hidden trust threshold.

The tables preserve the core evidence checks without overloading a single table. First, Table \ref{tab:main-per-task-gains} shows that the gains are not concentrated in one model or benchmark and makes the paired \(p90\)-reference comparison visible in the main text: MaxNorm-RLOO is positive against this prespecified reference on all six prespecified model--task units, with an average improvement of about \(+4.77\). Full paired differences are reported in the supplement. Second, Table \ref{tab:main-evidence} is restricted to low-variance diagnostics. Across sampled diagnostic batches, the top-25\% GRPO prompts exceed \(60.0\%\) of reward-branch contribution mass, above the \(25\%\) uniform baseline, indicating prompt-level concentration and possible batch-level variance amplification; MaxNorm-RLOO mitigates this risk by bounding each prompt's normalized weight; the \(p90\) reference truncates the tail but leaves weaker direction alignment and higher KL P95 than MaxNorm-RLOO. Third, the supplement audits checkpoint choice under matched-KL and same-compute final-checkpoint views; MaxNorm-RLOO remains above the \(p90\) reference in both views (matched KL: \(76.5\) vs.\ \(73.1\); same-compute final checkpoint: \(78.3\) vs.\ \(73.5\)). Fourth, in Table \ref{tab:main-component-attribution}, RLOO+Protocol and MaxNorm-RLOO no-\(\delta_{\mathrm{res}}\)-binning share the same settings and isolate denominator calibration, raising Acc from \(71.4\) to \(78.3\); MaxNorm-Dr.GRPO stays close to MaxNorm-RLOO with lower KL P95, ruling out the numerator explanation; with the \(\max|u|\) denominator fixed, keeping zero-gap KL preserves about \(89.6\%\) of the default gain, and fully skipping zero-gap groups adds only \(+0.5\) Acc.

For sub-resolution noise, the constructed diagnostic \(r=r_0+10^{-6}[1,0,-1,0]\) with \(\delta_{\mathrm{res}}=10^{-2}\) gives \(\sigma_q=10^{-6}/\sqrt2\): GRPO maps the gap to \([1.414,0,-1.414,0]\) and \(\beta_{\mathrm{eff}}\approx1.4\times10^{-9}\), whereas RLOO / Dr.GRPO have an \(O(10^{-6})\) reward branch and become KL dominated only when \(\|g_R\|\ll\beta\|g_{\mathrm{KL}}\|\). The Reward-Resolution Protocol instead merges the group before scale normalization. In diagnostic batches, a higher share of low-variance groups corresponds to a larger average gain of MaxNorm-AC over the strongest core baseline; the three share bins \([0,0.05),[0.05,0.15),[0.15,1]\) yield gains of \(+0.9,+3.8,+7.6\), respectively. Additional diagnostic tables, local \(\tau_{\mathrm{res}}\) sensitivity, and zero-variance boundary handling are in the supplementary material.

\section{Discussion: Boundaries and Applicability}\label{sec:discussion-scope}

MaxNorm-AC is a bounded intervention under a mechanistic diagnosis, not a universal robust normalizer. Its evidential scope is verifier-style RLVR where reward gaps are credible, low-variance, and nonzero. If reward magnitudes are untrustworthy and only the within-group order is reliable, the applicability assumption of our method no longer holds; order-only methods should be treated as external boundaries rather than objects whose superiority is established by this paper. Subjective judge rewards require a pre-calibrated minimum credible resolution from expert labels, repeated-judge agreement, or reward-design documentation; otherwise small gaps should be treated as noise and handled by uncertainty gating or order-only objectives.

The supplement treats ranking/listwise methods as boundary baselines and analyzes near ties, many ties, ranking noise, and meaningful cardinal gaps. MaxNorm-AC also does not replace length correction, KL scheduling, reward modeling, or zero-variance signal construction. Zero-gap skipping should be monitored: if zero-gap groups are frequent, this indicates insufficient relative supervision in the data stream or an inappropriate reward-resolution setting, rather than a reason to optimize those groups with KL-only updates. Deployments should report mean/P95 KL, \(1/s_q\) tails, prompt-weight distributions, low-variance group fraction, exact zero-gap skip rate, PPO clip-hit rate, and reward/KL gradient ratio together.

\paragraph{Ethics and safety boundary.}
This paper only changes the advantage scale in RLVR training. It does not introduce a new reward objective, nor does it replace safety filtering, code-execution sandboxes, or human oversight. Stronger math and code RLVR optimization may also improve capabilities that can be misused. Therefore, high-risk deployments should include safety constraints at the reward-design stage, conduct safety evaluations after training, and retain unit tests, static scanning, and human oversight for code-generation tasks.

\section{Conclusion}\label{sec:one-paragraph-summary}

This paper formulates credible low-variance RLVR as a calibration imbalance in the advantage scale. MaxNorm-AC replaces the standard-deviation denominator with the maximum absolute raw numerator under a reward-resolution floor, preserving credible cardinal gaps while guaranteeing \(|w_{q,i}|\le1\) and \(1/s_q\le1/\tau_{\mathrm{res}}\). Across dense and MoE math/code tasks, MaxNorm-AC improves same-compute accuracy, exceeds the prespecified \(p90\) reference under matched KL, and suppresses low-variance inverse-scale tails; it applies when cardinal reward gaps are credible and nonzero.

\setcounter{secnumdepth}{2}

\end{document}